\documentclass[copyright,creativecommons]{eptcs}
\usepackage{graphicx}
\usepackage{iftex}

\ifpdf
  \usepackage{underscore}         
  \usepackage[T1]{fontenc}        
\else
  \usepackage{breakurl}           
\fi

\title{Automation of Triangle Ruler-and-Compass Constructions Using Constraint Solvers}
\author{Milan Bankovi\' c
\institute{
  Faculty of Mathematics,
  University of Belgrade, Serbia\\
 }
\email{milan.bankovic@matf.bg.ac.rs}
}
\def\titlerunning{Automation of Triangle Ruler-and-Compass Constructions Using Constraint Solvers}
\def\authorrunning{Milan Bankovi\' c}
\begin{document}
\maketitle

\begin{abstract}
 In this paper, we present an approach to automated solving of triangle ruler-and-compass
 construction problems using finite-domain constraint solvers. The constraint model is described in the MiniZinc modeling language, and is based on the automated
 planning. The main benefit of using general constraint solvers for such
 purpose, instead of developing dedicated tools, is that we can rely on the efficient search that is already implemented within
 the solver, enabling us to focus on geometric aspects of the problem. We may also use the solver's built-in optimization capabilities to search for the shortest possible constructions. We evaluate our approach on 74 solvable problems from the Wernick's list, and compare it to the dedicated triangle construction solver ArgoTriCS. The results show that our approach is comparable to dedicated tools, while it requires much less effort to implement. Also, our model often
 finds shorter constructions, thanks to the optimization capabilities offered
 by the constraint solvers.
\end{abstract}

\section{Introduction}
\label{sect:introduction}

One of the oldest and the most studied classes of problems in geometry is the class of construction
problems: given some elements of a figure (such as a triangle), we want to find a sequence of steps to construct the remaining elements of the figure using the available tools -- typically a ruler\footnote{A more accurate term would be \emph{straightedge}, since a ruler is usually equipped with measuring marks, so it can be used to measure lengths, which is typically not allowed in geometric constructions. Nevertheless, in this paper we use the term \emph{ruler} and consider it as a synonym for a straightedge.} and a compass.
The beauty of this class of problems is that each problem is different and requires
a specific, often very deep geometric knowledge to be solved. Moreover, many problems
are even proven to be unsolvable.

Although geometricians love to deal with such problems by hand, for computer
scientists (who also love geometry) it is tempting to try to automate the solving
of construction problems. From the algorithmic point of view, the construction
problems are \emph{search} problems, and the search space is usually very large.
There are two main lines of approaches here: one is to develop a specific search
algorithm in some programming language with required geometric knowledge compiled
into it, and the other is to use existing artificial intelligence tools that are
good in solving search problems in general. In the second case, one should only
specify the problem and its constraints using some input language and then leave
the search to the tool.

In this paper, we advocate the second approach. More specifically, we show how
finite-domain constraint solvers \cite{rossi2006handbook} may be used for such purpose. We develop a
constraint model in the MiniZinc  modeling language \cite{nethercote2007minizinc}, based on the automated
planning \cite{ghallab2004automated}. There are two main benefits of using constraint solvers
for this purpose:

\begin{itemize}
 \item the constraint solvers are very efficient search engines, and by using
 them we may focus on geometric aspects of the problem and on modeling the
 geometric knowledge required for its solving, and leave the search
 to the tool that is good at it.
 \item the constraint solvers are usually equipped with optimization capabilities,
 enabling us to search for a construction that is the best in some sense (for instance,  the shortest possible construction may be required).
 This can be done with the minimal effort, compared to developing specific search algorithms with optimization capabilities (such as \emph{branch-and-bound} algorithms).
\end{itemize}

We compare our approach to the state-of-the-art tool for automated generation of triangle constructions ArgoTriCS \cite{marinkovic2017argotrics}, developed in the Prolog programming language. A
detailed evaluation is performed on 74 solvable problems from the Wernick's set
of triangle construction problems \cite{wernick1982triangle}.

The rest of this paper is organized as follows. In Section \ref{sect:background},
we introduce needed concepts and notation used in the rest of this paper. In Section \ref{sect:model} we describe our constraint model. Section \ref{sect:evaluation} provides a detailed evaluation of the approach. Finally,
in Section \ref{sect:conclusions}, we give some conclusions and mention some
directions of the further work.

\section{Background}
\label{sect:background}

\subsection{Ruler-and-Compass Constructions}

In this paper, we consider ruler-and-compass triangle constructions, where the
goal is to construct all vertices of a triangle, assuming that some elements
of the triangle (points, lines or angles) are given in advance. A construction
consists of a sequence of steps, where in each step some new objects (points, lines,
angles or circles) are constructed based on the objects constructed in previous
steps. Constructions performed in each of the steps are usually \emph{elementary} ones, such
as constructing the line passing through two given points, or the point that is the intersection of two given lines, or the circle centered at a given point that contains another given point. However, in
order to simplify the description of a triangle construction, some higher-level construction steps are also considered, such as constructing the tangents to a given circle from a given point, or the line perpendicular or parallel to a given line and passing through a given point, etc. Such higher-level constructions are called \emph{compound} constructions, since they can be easily decomposed into sequences of elementary construction steps.

In this paper, we focus on the Wernick's list of triangle construction problems \cite{wernick1982triangle}, where the following set of 16 characteristic points of a triangle is considered: the triangle vertices ($A$, $B$, $C$), the circumcenter $O$, the incenter $I$, the orthocenter $H$, the centroid $G$, the feet of the altitudes ($H_a$, $H_b$, $H_c$),
the feet of the internal angles bisectors ($T_a$, $T_b$, $T_c$) and the midpoints of
the triangle sides ($M_a$, $M_b$, $M_c$). Each problem from the list assumes that three different points from this set are given, and the goal is to construct all the
vertices of the triangle. There are 560 such point triplets, but only 139 among them represent significantly different problems (that is, mutually non-symmetric). Among these, only 74 problems are proven to be solvable by a ruler and a compass (others either contain
redundant points, or are undetermined, i.e.~may have
infinitely many solutions, or are proven to be unsolvable). In our work, we consider
only these 74 solvable problems from the Wernick's list.

In further text, we rely on the notation used by Marinkovi\' c \cite{marinkovic2017argotrics}. We also assume the geometric knowledge presented
in \cite{marinkovic2017argotrics}, as well as the set of elementary and compound
construction steps used in that work.

\subsection{Constraint Solving}

In this work, we reduce triangle construction problems to constraint solving \cite{rossi2006handbook}.
A finite-domain \emph{constraint satisfaction problem} (CSP) consists of a finite
set of \emph{variables} $\mathcal{X} = \{ x_1,\ldots, x_n \}$, each taking values from its given finite domain $D_i = D(x_i)$, and a finite set of \emph{constraints} $\mathcal{C} = \{ C_1,\ldots, C_m \}$, which are relations over subsets of these variables. A \emph{solution} of a CSP is an assignment $(x_1 = d_1, \ldots, x_n = d_n)$ of values to variables ($d_i \in D_i$), such that all
the constraints of that CSP are satisfied. A CSP is \emph{satisfiable} if it
has at least one solution, otherwise is \emph{unsatisfiable}. The optimization version of CSP, known as a \emph{constrained optimization problem} (COP) additionally assumes a function $f$ over the variables of the problem that should be minimized (or maximized), with respect to the constraints from $\mathcal{C}$.

Tools that implement procedures for solving CSPs (and COPs) are called \emph{constraint solvers}. They are usually based on a combination of
a backtrack-based search and constraint propagation \cite{rossi2006handbook}. Constraint solvers have been successfully used for solving many real-world problems
in many fields, such as scheduling, planning, timetabling, combinatorial design, and so on.

An important step in using constraint solvers is \emph{constraint modeling}, that is, representing a real-world problem in terms of variables and constraints. A constraint model is usually described using an appropriate \emph{modeling language}. One such
language supported by many modern constraint solvers is \emph{MiniZinc} \cite{nethercote2007minizinc}. This language offers a very flexible high-level
environment for modeling different kinds of constraints, enabling a compact and elegant way to represent some very complex problems. Examples of some high level
language elements include tuples, multi-dimensional arrays, sets, aggregate functions, finite quantification and so on. Since most of these high level constructs are not supported
by backend solvers, each MiniZinc model must be translated into an equivalent \emph{FlatZinc} form, containing only primitive language constructs and constraints
supported by a chosen backend solver. MiniZinc supports modeling of both CSPs and
COPs.

In MiniZinc, we distinguish \emph{variables} from \emph{parameters}. MiniZinc variables correspond to the variables of a CSP, i.e.~we declare their domains and expect from the solver to find their values
satisfying the constraints. On the other hand, parameters are just named constants, and their values must be known when the model is translated to the FlatZinc form (i.e.~before the solving starts). Parameters are the language's construct that allow us to specify a general model for a class of problems, and
then to choose a specific instance of the problem by fixing the values of the model's parameters. Parameter values are usually provided
in separate files (called \emph{data files}), so that we can easily combine the same model with different data.

In our work, we use MiniZinc as a modeling language.

\subsection{Automated Planning}

In our approach, triangle construction problems are considered as problems of \emph{automated planning} \cite{ghallab2004automated}. An automated planning problem consists of the following:

\begin{itemize}
 \item a set $\mathcal{S}$ of possible \emph{states}, which are usually encoded by a set of variables $\mathcal{V}$ and the values assigned to them. One distinguished state $S_0 \in \mathcal{S}$ is the \emph{initial state}.
 \item a set of operators $O$, where each operator $o \in O$ consists of a \emph{precondition} $C_o$ describing the conditions (in terms of the variables
 from $\mathcal{V}$) that must be satisfied in the current state for the operator to be applied, and
 a set of \emph{effects} $E_o$ (represented as variable-value assignments) describing how the current state is changed when $o$ is applied to it.
 The state obtained by applying an operator $o$ to some state $S$ is
 denoted by $o(S)$.
 \item a goal $G$, describing the conditions (in terms of the variables from $\mathcal{V}$) that must be satisfied in the final state.
\end{itemize}

The objective of automated planning is to find \emph{a plan}, that is, a finite sequence of operators $o_1,\ldots{},o_n$ from $O$ that can be successively
applied to the initial state
$S_0$ (i.e.~for each $i \in \{ 1,\ldots, n \}$, we have $S_i = o_i(S_{i-1})$, and the state $S_{i-1}$ satisfies the precondition $C_{o_i}$) producing the final state $S_n$ satisfying the goal $G$. The number $n$ of operators used in a plan is called the \emph{length} of the plan.

The problem of checking whether a plan (of any finite length) exists is  PSPACE-complete in general \cite{bylander1994computational}. For a fixed plan length $n$, the problem is NP-complete in general, and can be encoded as a CSP \cite{ghallab2004automated,rintanen2009planning}.

The optimization variant of the planning problem (i.e.~finding a plan of the minimal possible length) can be solved by successively checking for existence of plans
of lengths $n=1,2,3,\ldots$, that is, by solving the corresponding sequence of CSPs until a satisfiable one is encountered.

\section{Model Description}
\label{sect:model}

The triangle construction problems that we consider in this paper can be naturally
described as problems of automated planning:

\begin{itemize}
 \item states correspond to the sets of constructed objects, and the initial state is the set consisting of the given elements of the triangle (three points in case of Wernick's problems).
 \item operators correspond to the construction steps; the precondition for each operator is that objects used in the corresponding construction step are already
 constructed (i.e.~belong to the current state), and that corresponding non-degeneracy and determination conditions are satisfied (e.g.~two lines must be distinct and non-parallel in order to construct their intersection); the effect of each operator is the addition of the objects constructed by the corresponding construction step to the current state.
 \item the goal condition is that vertices $A$, $B$ and $C$ belong to the final
 state.
\end{itemize}

The corresponding planning problem for a fixed plan (construction) length is encoded as a CSP using the MiniZinc language.\footnote{The model
is available at: \url{https://github.com/milanbankovic/constructions/}.} In the rest of this section, we discuss different aspects of the encoding in more detail.

\subsection{Encoding of Geometric Knowledge}

\paragraph{Encoding objects.} We consider four types of objects: points, lines, circles and angles. Each of these types is encoded as an \emph{enumeration type}
in MiniZinc (\texttt{Point}, \texttt{Line}, \texttt{Circle} and \texttt{Angle}, respectively), and each object is represented by one enumerator of the corresponding type. The enumerated objects are the only objects that can be constructed. This means that we have to anticipate in advance the set of objects that might be needed during the construction.

\paragraph{Encoding relations.} Different relations between the enumerated objects are encoded by the parameters of the model, using MiniZinc's arrays, sets and tuples. These relations are used to statically encode the geometric knowledge used in the constructions. We define the following types of relations:

\begin{itemize}
 \item \emph{incidence relations}: we define two arrays of sets, \texttt{inc\_lines}
 and \texttt{inc\_circles}, indexed by points. The set \texttt{inc\_lines}[$p$] contains the lines incident with the point $p$, and the set \texttt{inc\_circles}[$p$]
 contains the circles incident with the point $p$.

 \item \emph{relations between lines}: we define two arrays of sets, \texttt{perp\_lines} and \texttt{paralell\_lines}, indexed by lines. The set \texttt{perp\_lines}[$l$] contains the lines perpendicular to the line $l$,
 and the set \texttt{parallel\_lines}[$l$] contains the lines parallel to the line $l$.

 \item \emph{circle tangents, diameters and centers}: the array of points \texttt{circle\_center} indexed by circles contains information about circle centers; the array \texttt{circle\_diameter} of point pairs indexed by circles contains information about circle diameters; the array \texttt{tangent\_lines} of
 sets of lines is indexed by circles, and the set \texttt{tangent\_lines}[$c$]
 contains the lines that are tangents of the circle $c$.

 \item \emph{vector ratios}: we use the array \texttt{known\_ratio\_triplets} of
 point triplets to store the information about the triplets of collinear points $(X,Y,Z)$ such that the ratio $\overrightarrow{XY}/\overrightarrow{YZ}$ is known. The exact value of the ratio is not encoded, since it is not important for the search (it is only important to know that we can construct one of the points if the remaining two are already constructed). Similarly, we use the array \texttt{known\_ratio\_quadruplets}
 to encode quadruplets of points $(X,Y,Z,W)$ such that the ratio $\overrightarrow{XY}/\overrightarrow{ZW}$ is known.

 \item \emph{angles between the lines}: we use the array \texttt{angle\_defs} of
 $\mathtt{Line} \times \mathtt{Line} \times \mathtt{Angle}$ triplets, to encode
 the information about the angles between the lines. A triplet $(p,q,\phi)$ means
 that the angle between the lines $p$ and $q$ is determined by $\phi$ (e.g.~is equal to $\phi/2$ or $\phi + \pi/2$). Such information can be used in two directions: if we have constructed $p$ and $q$, we can construct the angle $\phi$;
 also if we have constructed $p$ and $\phi$, and the intersection point $X$  of $p$ and $q$, we can then construct the line $q$.

 \item \emph{perpendicular bisectors of segments}: we use the array \texttt{perp\_bisectors} of $\mathtt{Point}\times\mathtt{Point}\times\mathtt{Line}$
 triplets to encode the information about the perpendicular bisectors of line segments.

 \item \emph{harmonic conjugates}: we use the array \texttt{harmonic\_quadruplets}
 of point quadruplets, where a quadruplet $(X,Y;Z,W)$ encodes that the points $X$ and $Y$ are harmonic conjugates of each other with respect to the pair $(Z,W)$.

 \item \emph{loci of points}: we use the array \texttt{locus\_defs} of $\mathtt{Point} \times \mathtt{Point} \times \mathtt{Angle} \times \mathtt{Circle}$
 tuples, where a tuple $(X,Y,\phi, c)$ encodes that the locus of
 points such that the segment $XY$ is seen at an angle determined by $\phi$ is an arc of the circle $c$.

 \item \emph{homothetic images of lines}: we use the array \texttt{homothety\_triplets}
 of $\mathtt{Point}\times\mathtt{Line}\times\mathtt{Line}$ triplets, where a triplet
 $(X,p,q)$ encodes that the line $q$ is the image of the line $p$
 by homothety centered in the point $X$ (again, homothety coefficient is not stored
 in the database).
\end{itemize}

\subsection{Encoding of the Planning Problem}

\paragraph{Encoding of states.} Let $n$ be the length of a plan that we are searching for, let $S_0$ be the initial state, and let $S_i$ be the state after the $i$th step.
To encode these states, we introduce arrays of variable sets \texttt{known\_points}, \texttt{known\_lines}, \texttt{known\_circles} and \texttt{known\_angles}, where, for instance, \texttt{known\_points}[$i$] ($i \in \{0,\ldots,n
\}$) denotes the set of points belonging to the state $S_i$ (similarly for other arrays). The initial state $S_0$
is fixed in advance by appropriate constraints (for instance $\mathtt{known\_points}[0] = \{ A, G, O \}$).

\paragraph{Encoding the plan.} We define the enumeration type \texttt{ConsType}, with one enumerator for each supported type of construction step. We also define the array \texttt{construct}  of variables of type \texttt{ConsType} (with indices in $\{1, \ldots, n \}$) encoding operators
used in each step (i.e.~the construction step types). For each step, we also need
additional information to fully determine the actual construction (for instance,
if we choose to construct the intersection of two lines, we must also choose the
lines that we want to intersect). For this reason, we also introduce additional
two-dimensional arrays of variables: for instance, \texttt{points}[$i$][$j$] denotes
the $j$th point used in the $i$th construction step (similarly we have \texttt{lines}[$i$][$j$], \texttt{circles}[$i$][$j$] and \texttt{angles}[$i$][$j$]).

\paragraph{Encoding the state transitions.} Finally, to glue the whole plan together,
we must add the constraints that connect the state variables in the successive
states, depending on the chosen operator in the corresponding step. This must
be done for each $i \in \{ 1, \ldots, n \}$, and that is where MiniZinc's finite
universal quantification comes in handy:

\begin{verbatim}
 constraint forall(i in 1..n)
 (
    construct[i] = LineIntersect ->
        % Precondition
        (lines[i,1] in known_lines[i-1] /\
         lines[i,2] in known_lines[i-1] /\
         lines[i,1] != lines[i,2] /\
         not (lines[i,1] in parallel_lines[lines[i,2]]) /\
         lines[i,1] in inc_lines[points[i,1]] /\
         lines[i,2] in inc_lines[points[i,1]] /\
         not (points[i,1] in known_points[i-1]) /\
        % Effects
         known_points[i] = known_points[i-1] union { points[i,1] } /\
         known_lines[i] = known_lines[i-1] /\
         known_circles[i] = known_circles[i-1] /\
         known_angles[i] = known_angles[i - 1]
        )
 );
\end{verbatim}
That is, for all $i \in \{1, \ldots, n\}$, if the chosen operator is $\mathtt{LineIntersect}$ (constructing the intersection of two lines), then
the chosen two lines $\mathtt{lines}[i,1]$ and $\mathtt{lines}[i,2]$ must
belong to the current state $S_{i-1}$ (i.e.~they must have been already constructed),
they must be distinct and not parallel. Also, the chosen point $\mathtt{points}[i,1]$
must belong to both chosen lines (i.e.~it must be their intersection), and it must not belong to the current state (we do not want to construct a point that is
already constructed). If all these preconditions are met, then the effect is
that the set $\mathtt{known\_points}[i]$ is obtained by adding the intersection point $\mathtt{points}[i,1]$ to the set $\mathtt{known\_points}[i-1]$ (the sets of lines, circles and angles remain the same). Similar constraints are defined for all other types of construction steps.

\paragraph{Encoding the goal.} The goal is encoded simply by adding the constraints that require that the vertices $A$, $B$ and $C$ belong to the set $\mathtt{known\_points}[n]$:

\begin{verbatim}
 { A, B, C } subset known_points[n];
\end{verbatim}

\section{Evaluation}
\label{sect:evaluation}

The model described in the previous section is evaluated on 74 solvable instances
from Wernick's set \cite{wernick1982triangle}. The experiments were performed on
a computer with 3.1GHz processor and 8Gb of RAM. We used official MiniZinc distribution\footnote{\url{https://www.minizinc.org/software.html}} for experiments (version 2.7.2).
We have experimented with different backend constraint solvers provided within
MiniZinc distribution, and by far the best results were obtained by the \texttt{chuffed}\footnote{\url{https://github.com/chuffed/chuffed}}
solver. Therefore, in the rest of this section, we present only the results obtained by \texttt{chuffed}.

We looked for plans of minimal lengths (i.e.~constructions with the minimal possible numbers of steps). We used three different setups:

\begin{itemize}
 \item \emph{linear setup}: for each of the problems, we successively look for plans of length $n=1,2,3,\ldots$, and stop when we encounter a satisfiable CSP, or when some upper limit $maxSteps$ is exceeded. This is the usual way for finding plans of minimal lengths in automated planning \cite{rintanen2009planning}. In our experiments, the upper limit for the plan length was set to 11, since our preliminary experiments had shown that all the problems that our model could solve had been solved in at most 11 steps. Note that in this setup the value of $maxSteps$ does not affect the solving time for problems that our model can solve (that is, using a greater value of $maxSteps$ would not slow down
 the search).

 \item \emph{minimization setup}: we reformulate our model such that the plan length $n$ is not fixed. Instead, $n$ is a variable with a domain $\{1,\ldots, maxSteps\}$ and
 we are trying to minimize the value of $n$ (that is, we are solving a constrained optimization problem). The problem with this approach is how to determine the value
 of $maxSteps$ parameter, since in this setup greater values of this parameter make the model larger and the search becomes slower, even for problems that can be solved in a small number of steps. In our experiments, we used the value $maxSteps=11$, but this was somewhat artificial choice, since we used the previous
 knowledge to choose the minimal possible number of steps sufficient to solve all
 the problems that our model was able to solve.

 \item \emph{incremental setup}: just like in the previous setup, we reformulate our model such that we are trying to minimize $n$, but this time the domain for $n$ is some interval $\{l,\ldots,u\}$, where $l$ and $u$ are parameters. Now we successively solve constrained optimization problems for intervals $\{1,\ldots,k\}$, $\{ k+1,\ldots,2k \}$,$\{ 2k+1,\ldots, 3k \}$,\ldots, until some of them turns out to be satisfiable, or until some upper limit $maxSteps$ is exceeded. Like in the first setup, the choice for the value of the parameter $maxSteps$ does not affect the solving time for the problems that are solvable by our model. On the other hand, the number of COPs solved is smaller roughly by the factor $k$, compared to the first setup. We have experimented with multiple
 choices for $k$, and the best results were obtained for $k=3$.

\end{itemize}

In Table \ref{tab:results}, we provide the main results of our evaluation. We have evaluated all
three setups described above. We also compared our approach to the results obtained
by the ArgoTriCS dedicated triangle construction solver developed by Marinkovi\' c \cite{marinkovic2017argotrics}.
ArgoTriCS is implemented in Prolog programming language, but it uses a very similar
knowledge base and an almost identical set of available construction steps.

\begin{table}[!ht]
\begin{center}
{\footnotesize
\begin{tabular}{cccccc}
 \textbf{Setup} & \textbf{\# solved} & \textbf{Avg.~time} & \textbf{Median time} & \textbf{Avg.~time on solved} & \textbf{Avg.~length} \\
 \hline
 linear & 63 & 97.9 & 22.0 & 58.5 & 6.3 \\
 minimization & 63 & 43.8 & 10.8 & 29.7 & 6.3 \\
 incremental ($k=3$) & 63 & 66.1 & 12.0 & 39.9 & 6.3 \\
 \hline
 ArgoTriCS  & 65 & 54.5 & 21.6 & 54.4 & 7.5 \\
 \hline
\end{tabular}

}
\end{center}
 \caption{Overall results for different setups, compared to ArgoTriCS. Times are given in seconds}
 \label{tab:results}
\end{table}

Note that the choice of the setup does not affect how many problems from Wernick's list will be solved, since this depends only on the geometric knowledge that is
compiled into our model.\footnote{This means that we can improve our results by carefully examining the knowledge needed for solving the unsolved problems, and incorporating that knowledge into our model. However, such enrichment of the model enlarges the search space and makes the solving slower even for the problems that are already solvable by our model.} In total, we managed to solve 63 of 74 problems (for the remaining 11 problems, the constraint solver reported unsatisfiability).
On the other hand, ArgoTriCS solved 2 problems more. This is because we missed
to incorporate some of the objects and lemmas known to ArgoTriCS to our model.

The best average solving time is obtained by the minimization setup. However,
as we mentioned earlier, the average solving time in this setup greatly depends
on the choice for the maximal possible value of $n$. The results shown in Table
\ref{tab:results} are obtained for $maxSteps=11$. We also experimented with
some greater values. For instance, for $maxSteps=20$ the average solving time
was over 100 seconds, that is, more than twice greater (of course, the number of solved problems remained the same).

The linear setup has shown the worst performance. This is because in this setup
we were solving many unsatisfiable CSPs until we possibly reached some satisfiable
CSP. Unsatisfiable CSPs tend to consume more time, especially those that are
``almost satisfiable'', that is, that are close to some phase transition point.
This phenomenon is well-known in automated planning \cite{rintanen2009planning}.

\begin{figure}[!ht]
\begin{center}
\includegraphics[width=460pt]{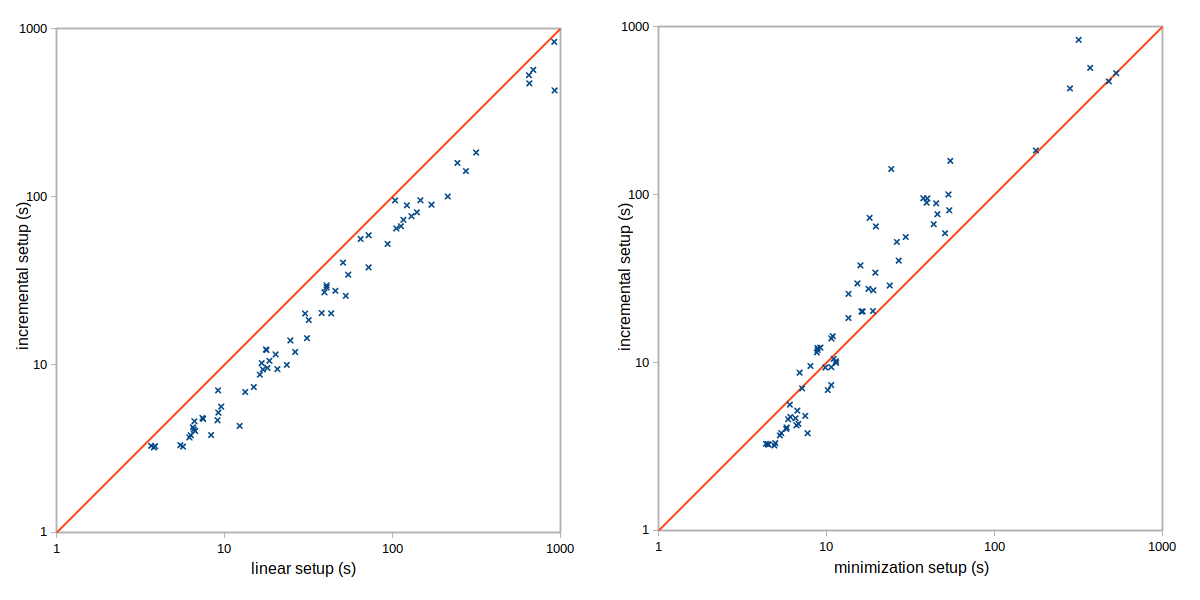}
\end{center}
\caption{Per-instance comparison of different setups. Times are given in seconds}
\label{fcs}
\end{figure}

The performance of the incremental setup was much better on average than in case of the linear setup, and a little worse than in case of the minimization setup, but still comparable. A more detailed, per-instance comparison is shown in Figure \ref{fcs}. We can see that the incremental setup was uniformly better than linear setup, and was also better than the minimization setup on easier instances, but it was outperformed by the minimization setup on harder problems.
Overall, the incremental setup seems as a good choice in a realistic context, when we do not know in advance the value of $maxSteps$ parameter.

\begin{figure}[!ht]
\begin{center}
\includegraphics[width=300pt]{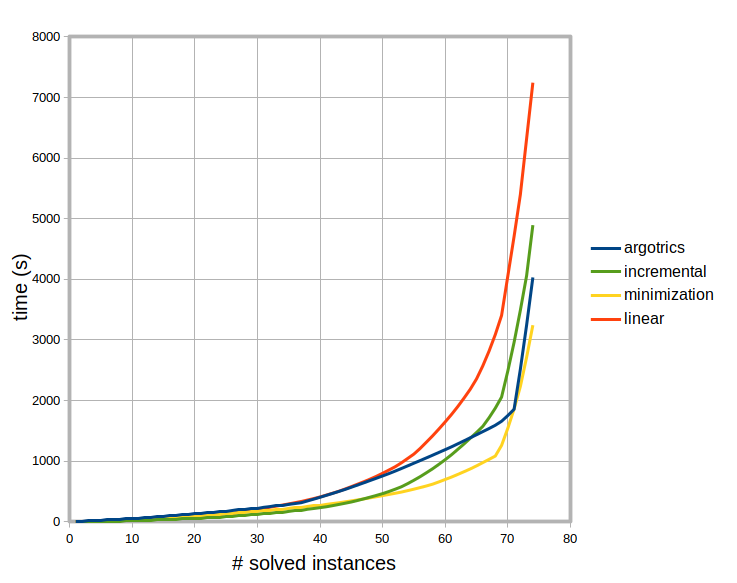}
\end{center}
\caption{Survival plot for all three setups, compared to ArgoTriCS. Times are given in seconds}
\label{fs}
\end{figure}

The overall performance of the ArgoTriCS solver was comparable to our
approach, when the average solving time is concerned. However, we may notice
that its median solving time was almost twice greater than in case of our
minimization or incremental setup. Also, the average solving time on solved
instances was much better in our approach. This suggests that our approach
performed better than (or comparable to) ArgoTriCS on problems for which it managed to find a construction plan, especially on easier instances. This is confirmed in Figure \ref{fs}, which shows the survival plot for all three setups and ArgoTriCS. The minimization setup was clearly the best, while the linear setup was the worst. When compared to ArgoTriCS, the incremental setup was cumulatively better on more than 60 instances, which were roughly all the instances that our model managed to solve.
This means that if our model can find a solution, it can do it fast, while its performance is much worse when it comes to the instances that are out of its reach (that is, when the corresponding CSPs are unsatisfiable). On the other hand, the performance of ArgoTriCS had much smaller variance -- it performed almost equally solid on all instances (as it can be seen from Table \ref{tab:results}, the average solving time on solved instances for ArgoTriCS is almost the same as the average solving time on all instances).

\begin{figure}[!ht]
\begin{center}
\includegraphics[width=250pt]{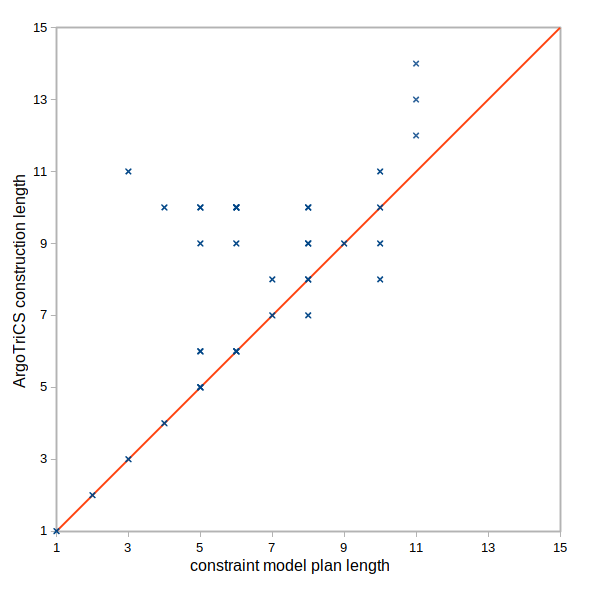}
\end{center}
\caption{A per-instance comparison of construction (plan) lengths between ArgoTriCS and our approach}
\label{fpl}
\end{figure}

The final comparison between ArgoTriCS and our approach concerns the lengths of the obtained constructions. Table \ref{tab:results} shows that the average plan length
in our approach was 6.3 (again, this does not depend on the chosen setup). On the other hand, the average number of steps in ArgoTriCS's constructions was 7.5. Notice
that these numbers are comparable, since the sets of available construction steps
in both systems are almost identical. A more detailed, per-instance comparison is shown in Figure \ref{fpl}. The plot clearly confirms that our approach is by far superior when finding the shortest constructions is concerned. However, for the sake of fairness, we must stress that ArgoTriCS was not designed with that optimization in mind, that is, it does not even search for the shortest solutions.
We guess that such a capability could be integrated in ArgoTriCS, but with much
more effort, since it would have to be manually implemented in Prolog (just like the
search itself). On the other hand, in our approach, we rely on the built-in capabilities of constraint solvers to solve optimization problems efficiently,
imposing the minimal possible effort on our side.

\section{Conclusions and Further Work}
\label{sect:conclusions}

In this paper we presented and evaluated a method for automated triangle
construction based on constraint solving. We compared our method to the
state-of-the-art dedicated triangle construction solver ArgoTriCS, developed in Prolog programming language. We advocate that our approach has two important advantages. First,
our approach is much simpler to implement, since we rely on powerful constraint
solvers which can efficiently do the search for us, and we may focus only on modeling. On the other side, in the ArgoTriCS solver the search is implemented by hand, in more than 500 lines of code. Second, we can easily employ the optimization capabilities of
modern constraint solvers to search for the shortest possible constructions, while implementing such functionality in ArgoTriCS would require much
more effort.

We evaluated our approach on 74 solvable problems from the Wernick's list. The results
showed that our approach is comparable to ArgoTriCS when solving time is concerned.
On the other hand, our model often finds shorter constructions, due to built-in
optimization capability which is missing in ArgoTriCS.

For further work, we plan to extend our model to support construction problems
from other sets. This should not be a hard task in the technical sense, since the
model is developed such that it can be easily extended (that is, we can easily add new objects, relations and construction step types). The real challenge is to recognize
and integrate the geometric knowledge needed for such constructions into the model. Of course, this
is a job for geometricians, and our goal was to provide them with (what we hope is) a useful tool that can
free them from the tedious task of programming, and let them focus on what they do the
best and love the most.

\paragraph{Acknowledgements.}
This work was partially supported by the Serbian Ministry of
  Science grant 174021. We are very grateful to the anonymous reviewers
  whose insightful comments and remarks helped us to make this paper much
  better.

\nocite{*}
\label{sect:bib}
\bibliographystyle{eptcs}
\bibliography{references}

\end{document}